\documentclass[11pt]{article}

\usepackage[final]{acl}

\usepackage{times}
\usepackage{latexsym}
\usepackage{booktabs}
\usepackage{multirow}
\usepackage{algorithm}
\usepackage{algorithmic}
\usepackage{xcolor}
\usepackage{tipa} 
\usepackage{pifont}

\usepackage[T1]{fontenc}

\usepackage[utf8]{inputenc}

\usepackage[arabic,french, english]{babel}
\usepackage{CJKutf8}
\usepackage{microtype}

\usepackage{inconsolata}

\usepackage{graphicx}

\newcommand{\huggingfacesmall}{\includegraphics[width=9px]{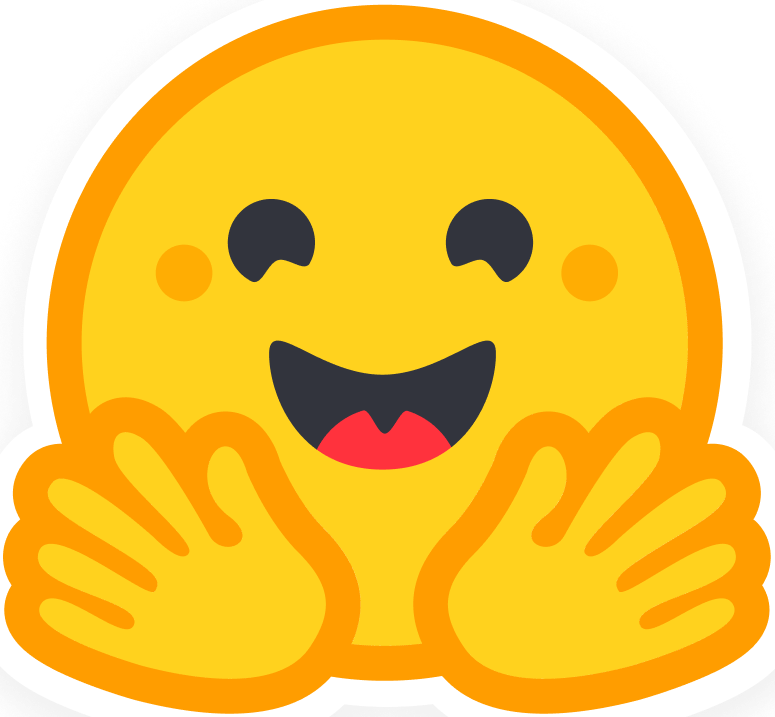}}
\newcommand{\githubsmall}{\includegraphics[width=10px]{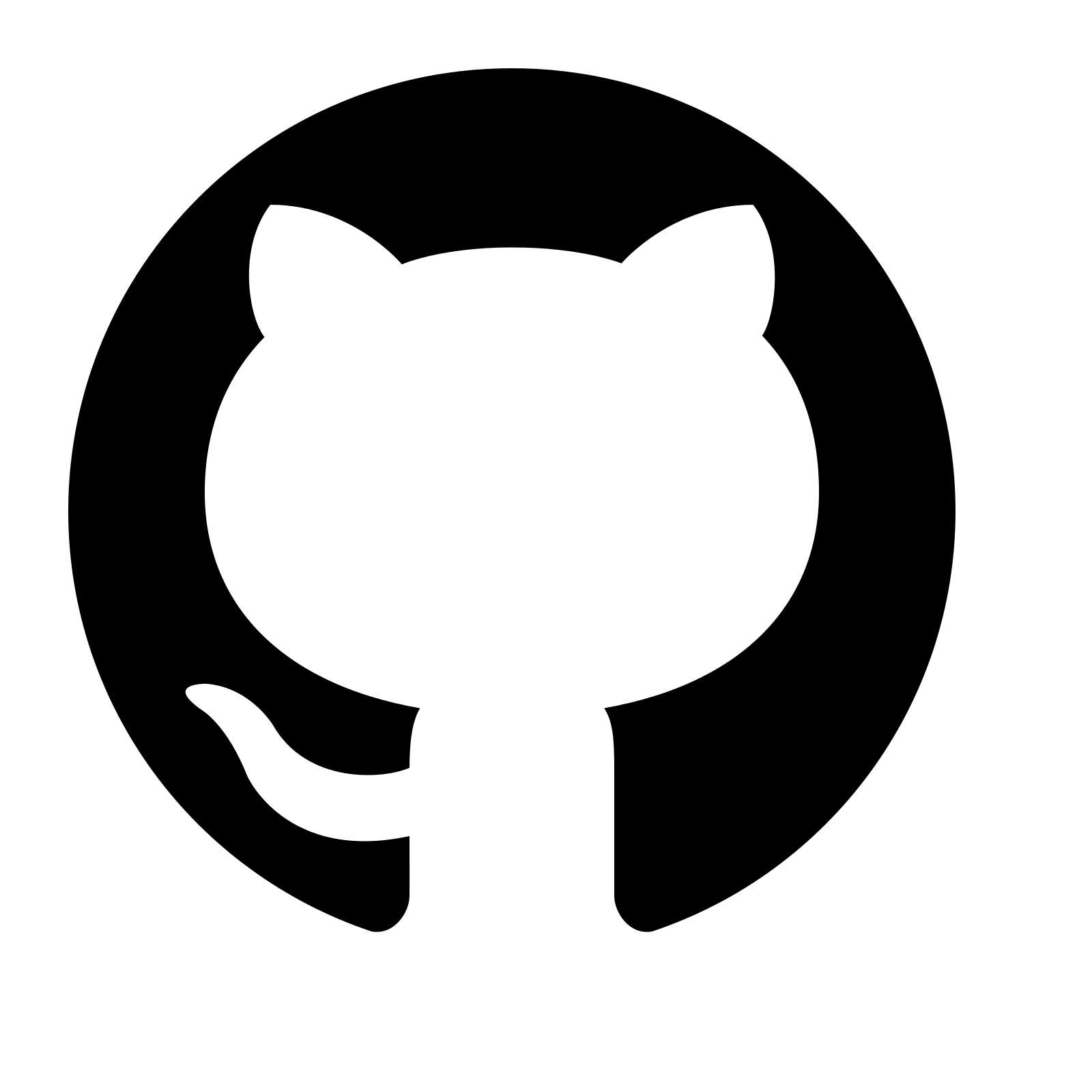}}

\title{KIT's Submission to Cross-Lingual Voice Cloning in IWSLT 2026}

\author{
  \textbf{Seymanur Akti\textsuperscript{1,3}},
  \textbf{Alexander Waibel\textsuperscript{1,2}},
\\
  \textsuperscript{1} Karlsruhe Institute of Technology (KIT)\\
  \textsuperscript{2} Carnegie Mellon University (CMU),
  \textsuperscript{3} KIT Campus Transfer (KCT)
\\
  \small{
    \textbf{Correspondence:} \href{mailto:seymanur.akti@kit.edu}{seymanur.akti@kit.edu}
  }
}

\begin{document}
\maketitle
\begin{abstract}
Cross-lingual voice cloning aims to generate speech in a target language while preserving speaker identity from a source-language reference. This task is central to speech translation and is the focus of the IWSLT 2026 Cross-Lingual Voice Cloning track. A key challenge is maintaining intelligibility and naturalness in the presence of accent variation and domain-specific vocabulary. We build on a multilingual text-to-speech model, FishAudio-S2-Pro, and introduce language tag prompting to improve language control and reduce accent leakage. We further apply reinforcement learning (RL) fine-tuning for task adaptation and observe improvements in intelligibility. Finally, we propose a reference-conditioned lexical matching method that improves pronunciation of domain-specific terms when lexical overlap is present.

Results show that language prompting provides the largest gains, while lexical matching yields consistent improvements on matched subsets.

\end{abstract}
\section{Introduction}

Recent advances in multilingual text-to-speech (TTS) systems have enabled high-quality speech synthesis across multiple languages within a single model, while also supporting in-context voice cloning from reference audio. This capability is particularly important for speech translation pipelines, where cross-lingual voice cloning aims to generate speech in a target language while preserving speaker identity and speaking style from a source-language reference.

In IWSLT 2026~\cite{adelani-etal-2026-iwslt}, Cross-Lingual Voice Cloning track focuses on this setting, requiring synthesis in French, Arabic, and Chinese from English reference speech. The dataset includes diverse speaker accents and domain-specific terminology, which increases the difficulty of maintaining pronunciation consistency under cross-lingual conditions.

Cross-lingual voice cloning can be formulated using cascaded pipelines, where speech is first synthesized using a TTS system and then transformed using a separate voice conversion model. While such cascaded approaches offer modularity and flexibility, they introduce additional sources of error, including speaker leakage~\cite{akti2025towards} and misalignment between linguistic content and prosody.

An alternative direction is end-to-end conditioning of speech synthesis on speaker characteristics. One line of work incorporates speaker information via explicit speaker encoders, which inject speaker representations into the synthesis model~\cite{casanova2022yourtts, lee2025hierspeech++, li2023styletts}. This approach remains widely used due to its simplicity and stability; however, its voice cloning capability is constrained by the quality and expressiveness of the learned speaker embeddings.

More recent work has explored in-context learning approaches for zero-shot voice cloning, where models are conditioned directly on reference audio to generate speech across unseen speakers and languages. Systems such as VALL-E~\cite{wang2023neural}, F5-TTS~\cite{chen2025f5}, Qwen3-TTS~\cite{hu2026qwen3}, and CosyVoice3~\cite{du2025cosyvoice} follow this paradigm and demonstrate strong performance in preserving speaker identity in zero-shot settings. However, cross-lingual generation remains challenging even for large-scale models. A common issue is accent leakage, where phonetic characteristics of the source language persist in the generated speech. In addition, domain-specific words and named entities are often mispronounced, especially when they are out-of-distribution for the target language.

In this work, we present Karlsruhe Institute of Technology's (KIT) submission to the IWSLT 2026 Cross-Lingual Voice Cloning track. We build on a strong pretrained multilingual TTS model, FishAudio-S2-Pro\footnote{\githubsmall \href{https://github.com/fishaudio/fish-speech}{fishaudio/fish-speech}}, and focus on improving cross-lingual pronunciation through input-level conditioning and lightweight adaptation strategies.

\section{Data}

We evaluate our system on the ACL 60/60 dataset~\cite{salesky2023evaluating}\footnote{\huggingfacesmall \href{https://huggingface.co/datasets/ymoslem/acl-6060}{ymoslem/acl-6060}}, which consists of long-form speech recordings derived from ACL conference talks. The dataset is designed for cross-lingual voice cloning and follows the IWSLT 2026 shared task setup, where English reference speech is paired with target text in French, Arabic, and Chinese.

The dataset presents several challenges. First, reference speech exhibits substantial variability in speaker accents, which introduces inconsistencies in speaker representation and complicates cross-lingual transfer. Second, the target text contains domain-specific terminology and named entities, many of which are either out-of-distribution for the target languages or appear in their original English form. This makes accurate pronunciation particularly challenging under cross-lingual conditions.

We use the ACL 60/60 development set for system development and reinforcement learning fine-tuning, and reserve the evaluation set for validation and analysis. All recordings are provided with gold utterance-level segmentation, and no additional segmentation or alignment is performed during fine-tuning. For each target text sample, the corresponding English reference speech-text pairs are used for conditioning.

In the blind test scenario, reference audio is provided in long-form format. To enable effective conditioning, we use an ASR model to segment the audio into smaller chunks around 2 to 10 seconds paired with their corresponding text transcriptions. We then perform inference-time retrieval over these segments when required by our lexical matching strategy. This retrieval is applied only at inference time and does not introduce additional supervision.

\section{Model Adaptation}

We build our system on top of FishAudio-S2-Pro, a multilingual text-to-speech model that supports in-context voice cloning from reference audio~\cite{liao2026fish}. The model is capable of synthesizing speech in multiple languages, including Arabic, French, and Chinese, making it suitable for the IWSLT cross-lingual voice cloning task.

FishAudio-S2-Pro allows flexible text prompting, enabling the use of free-form control tokens in the input. In our work, we leverage this capability to introduce explicit language tags for improved language control during generation. To simulate a realistic scenario where target speech is unavailable, we perform reinforcement learning fine-tuning instead of supervised training, aiming to maintain the original model quality while improving perceptual performance in cross-lingual settings.

\subsection{Prompts with Language Tags}

FishAudio-S2-Pro does not explicitly use language tags during training, and language identification is implicitly inferred from the input text. In cross-lingual voice cloning with autoregressive generation over mixed-language sequences, this can lead to accent leakage when multiple languages are present within a sequence.

To address this, we leverage the model’s support for free-form prompting and introduce explicit language tags for each text input, including the reference text. We experiment with both English-language tags (e.g., [english], [arabic], [french], [chinese]) and native-script tags (e.g., [english], [\AR{العربية}], [\foreignlanguage{french}{français}], \begin{CJK*}{UTF8}{gbsn}{[普通话]}\end{CJK*}). We find that native-script tags provide stronger conditioning signals and yield better cross-lingual pronunciation quality in our setup.

Overall, language tags act as a simple but effective control mechanism, guiding the model toward the desired phonetic realization and reducing cross-lingual interference.

An example of the constructed prompt for cross-lingual generation for submission is shown below:

\begin{quote}
\textbf{Reference:} \\
\texttt{\textcolor{blue}{[english]} The little cat is sleeping under the table.}

\vspace{4pt}

\textbf{Chinese (Mandarin) Target:} \\
\texttt{\begin{CJK*}{UTF8}{gbsn}{\textcolor{blue}{[普通话]} 小猫正在桌子下面睡觉。}\end{CJK*}}

\vspace{8pt}

\textbf{French Target:} \\
\texttt{\textcolor{blue}{[\foreignlanguage{french}{français}]} Le petit chat dort sous la table.}

\vspace{8pt}

\textbf{Arabic Target:} \\
\texttt{\textcolor{blue}{[\AR{العربية}]} \AR{القطة الصغيرة نائمة تحت الطاولة.}}

\end{quote}
\subsection{RL Fine-tuning}

To adapt the model to the cross-lingual voice cloning task, we perform reinforcement learning (RL) fine-tuning using Group Relative Policy Optimization (GRPO)~\cite{shao2024deepseekmath}, which has already shown effectiveness in improving the base model. The objective is to adapt the model to the newly introduced language tags and cross-lingual inference without requiring supervised parallel data.

We define the reward function based on two components: (1) Character Error Rate (CER), computed using a multilingual ASR model~\cite{pratap2024scaling}\footnote{\huggingfacesmall \href{https://huggingface.co/facebook/mms-1b-all}{facebook/mms-1b-all}}, and (2) speaker similarity (SSIM), computed using a speaker verification model~\cite{chen2022wavlm}.\footnote{\huggingfacesmall \href{https://huggingface.co/microsoft/wavlm-base-plus-sv}{microsoft/wavlm-base-plus-sv}}

\subsubsection{Hyperparameters and Implementation Details}

During RL fine-tuning, we adopt a more aggressive update strategy by optimizing attention, MLP, and output layers, compared to the original recipe which only updates MLP layers. We use LoRA with rank $r=64$ and scaling factor $\alpha=16$. The group size for GRPO was chosen as 8 and we use AdamW optimizer with learning rate $10^-5$

In addition, we experiment with different values of the KL divergence penalty coefficient, which controls deviation from the base model. Proper tuning of this coefficient improves training stability and leads to better adaptation to the target task. We used $\beta=0.1$ as the penalty coefficient.

For reward calculation, the CER score is inverted, and both CER and SSIM are scaled to [0,1]. The final reward is computed as their average:

\begin{equation}
    {Reward} = \frac{(1 - {CER}) + {SSIM}}{2}
\end{equation}

\subsection{Reference Speech Retrieval from Long Audio}

To better handle domain-specific vocabulary in the blind test set, we introduce a reference speech retrieval strategy from long-form audio. Given long reference recordings, we first segment the audio into smaller chunks and transcribe them using the VibeVoice long-form ASR model~\cite{peng2026vibevoice}.\footnote{\huggingfacesmall \href{https://huggingface.co/microsoft/VibeVoice-ASR}{microsoft/VibeVoice-ASR}}

We then perform lexical matching between the target text and transcribed audio segments, selecting reference segments that contain overlapping words. This allows the model to observe correct pronunciations of domain-specific terms and rare words directly from the retrieved reference speech.

By conditioning on these retrieved acoustic segments, this approach improves pronunciation accuracy while preserving speaker-specific characteristics.

\section{Evaluation Results}

We evaluate our model on the ACL 60/60 evaluation subset, covering Arabic, Chinese, and French as target languages. Each sample is synthesized in all three languages given English reference speech. We compare three settings: (i) a baseline model (FishAudio-S2-Pro) without language tags, (ii) the same model with explicit language tags, and (iii) an RL fine-tuned model using language-tagged cross-lingual prompts.

We use the following evaluation metrics:
\begin{itemize}
    \item \textbf{CER}: Character Error Rate computed using Whisper-Large-v3~\cite{radford2023robust}\footnote{\huggingfacesmall \href{https://huggingface.co/openai/whisper-large-v3}{openai/whisper-large-v3}} for intelligibility assessment. We use CER instead of WER to ensure comparability across languages.
    
    \item \textbf{SSIM}: Speaker similarity calculated by cosine similarity between speaker embeddings extracted from source and generated speech using a pre-trained speaker verification model~\cite{speechbrain}.\footnote{\huggingfacesmall \href{https://huggingface.co/speechbrain/spkrec-ecapa-voxceleb}{speechbrain/spkrec-ecapa-voxceleb}}
    
    \item \textbf{UTMOS}: Predicted mean opinion score using a pre-trained MOS estimation model~\cite{saeki2022utmos}.
\end{itemize}

We use different ASR and speaker verification models from those used in the reward computation during RL fine-tuning to avoid evaluation bias and ensure a fair assessment of generalization performance.

\renewcommand{\arraystretch}{1.15}

\begin{table*}[t]
\centering

\setlength{\tabcolsep}{4pt}

\begin{tabular}{lc ccc ccc ccc}
\toprule
& & \multicolumn{3}{c}{\textbf{CER (\%) $\downarrow$}} 
& \multicolumn{3}{c}{\textbf{SSIM (\%) $\uparrow$}} 
& \multicolumn{3}{c}{\textbf{UTMOS $\uparrow$}} \\
\cmidrule(lr){3-5} \cmidrule(lr){6-8} \cmidrule(lr){9-11}

\textbf{Model} & \textbf{Lang. Tags}
& \textbf{ar} & \textbf{fr} & \textbf{zh} 
& \textbf{ar} & \textbf{fr} & \textbf{zh} 
& \textbf{ar} & \textbf{fr} & \textbf{zh} \\

\midrule

Baseline & \ding{55}
& 6.57 & 3.10 & 11.37 
& 64.05 & 60.75 & 62.39 
& \textbf{2.94} & 2.86 & \textbf{2.90} \\

Baseline & \ding{51} 
& 6.39 & 2.90 & 12.05 
& 63.77 & 60.21 & 62.49 
& \textbf{2.94} & 2.85 & 2.88 \\

RL Finetuned $^\dagger$ & \ding{51}
& \textbf{6.38} & \textbf{2.78} & \textbf{10.99} 
& \textbf{64.15} & \textbf{60.83} & \textbf{62.52} 
& 2.93 & \textbf{2.88} & 2.89 \\

\bottomrule
\end{tabular}

\caption{Evaluation across languages. CER is reported in \% (lower is better), while speaker similarity and UTMOS are higher-is-better metrics. $^\dagger$ indicates submitted system.}
\label{tab:main_results_clean}
\end{table*}

The results in Table~\ref{tab:main_results_clean} show that introducing language tags on the base model improves CER for Arabic and French, while a slight degradation is observed for Chinese. This suggests that language conditioning might help reduce cross-lingual pronunciation drift even when it is used without prior tuning, although its effect is inconsistent across languages.

For speaker similarity, adding language tags leads to only minor variations, indicating that linguistic conditioning primarily affects phonetic realization while preserving speaker identity.

After RL fine-tuning with the language tags, we observe small but consistent improvements or stability across all languages. In particular, CER improves further compared to the language-tag-only setting, indicating that RL fine-tuning further refines pronunciation consistency under cross-lingual conditioning. Speaker similarity and UTMOS remain stable, suggesting that RL fine-tuning does not degrade speaker identity or perceptual quality.

\subsection{Source Language Bias Analysis}

We further find that language tags primarily reduce English pronunciation bias, leading to more natural target-language realization. Without language tags, generated speech often retains English-influenced phonetic characteristics, particularly in long-form reference conditions, which can amplify this bias during autoregressive generation. This effect is reduced when language tags are introduced, resulting in more consistent target-language pronunciation and prosody.

To evaluate this behaviour, we use a pre-trained language identification model\footnote{\huggingfacesmall \href{https://huggingface.co/speechbrain/lang-id-voxlingua107-ecapa}{speechbrain/lang-id-voxlingua107-ecapa}} to measure target language confidence in generated speech. We hypothesize that leakage from the reference speech introduces pronunciation drift, which can negatively affect language identification performance.

Table~\ref{tab:lid} shows that language tags consistently improve target language identification probabilities across all languages, indicating reduced cross-lingual pronunciation leakage and improved language consistency.

At the same time, speaker-specific characteristics remain largely preserved, including speaking rate, accent-related phonetic traits, and speaking style. For example, speakers with strong regional accents tend to retain similar accent characteristics across target languages, indicating that language conditioning primarily affects linguistic realization rather than speaker identity.

\begin{table}[t]

\centering
\small
\renewcommand{\arraystretch}{1.3}

\begin{tabular}{lcccc}

\toprule
\textbf{Model} & \textbf{ar} & \textbf{fr} & \textbf{zh} & \textbf{avg} \\
\midrule

Normal Prompts & 89.87 & 88.68 & 90.99 & 89.85 \\
w/ Language Tags & \textbf{93.42} & \textbf{90.23} & \textbf{92.13} & \textbf{91.64} \\

\bottomrule
\end{tabular}

\caption{Target language identification probabilities (\%) with and without explicit language tags. Language conditioning improves target language confidence across all languages.}
\label{tab:lid}
\end{table}

\subsection{Impact of Lexical Matching on Special Word Pronunciation}

We further analyze pronunciation behavior on named entities and domain-specific terms under matched and non-matched reference conditioning. The matched setting is constructed using reference audio corresponding to the exact translation of the target text, such that domain-specific terms are already present in the reference prompt and provide explicit phonetic grounding. In contrast, the non-matched setting uses randomly selected reference audio from the same speaker, which does not necessarily contain the target lexical content.

As shown in Table~\ref{tab:special_words}, matched reference audio significantly improves pronunciation consistency for out-of-language words and acronyms. In this setting, the model is able to better preserve correct pronunciations when the reference speech provides aligned lexical and phonetic cues. We also observe that speaker-specific pronunciation patterns are preserved for named entities, reflecting the way the speaker produces these terms in the reference audio.

In contrast, under non-matched conditions, the model exhibits weaker phonetic grounding and tends to rely on spelling-based or language-default pronunciation strategies. This often results in letter-by-letter or segmented pronunciation of technical terms and acronyms. For example, tokens such as “seq2seq” or “word2vec” are interpreted compositionally, where the digit “2” is pronounced as a number rather than as part of the unit. This highlights the importance of reference-conditioned lexical grounding for accurate pronunciation.

\begin{table}[t]
\centering
\small
\renewcommand{\arraystretch}{1.4}

\begin{tabular}{lcc}
\toprule
\textbf{Entity} & \textbf{Matching} & \textbf{Non-matching} \\
\midrule

VALSE 
& \textcolor{green!60!black}{\textipa{/v{A}ls/}}
& \textipa{\textcolor{red!70!black}{/vi: eI El Es i:/}} \\

LXMert 
& \textipa{/El Eks \textcolor{green!60!black}{m3:rt}/} 
& \textipa{/El Eks \textcolor{red!70!black}{Em 3:rt}/} \\

ViLBERT 
& \textipa{/\textcolor{green!60!black}{vIl}b3:rt/}
& \textipa{/\textcolor{red!70!black}{vi: El} b3:rt/} \\

Word2Vec 
& \textipa{/w3:d tu \textcolor{green!60!black}{vEk}/} 
& \textipa{/w3:d tu  \textcolor{red!70!black}{vIk}/} \\

RNSum 
& \textipa{/A:r En \textcolor{green!60!black}{sVm}/}
& \textipa{A:r En \textcolor{red!70!black}{Es Vm}/} \\

SVAMP 
& \textcolor{green!60!black}{\textipa{/swA:mp/}} 
& \textcolor{red!70!black}{\textipa{/Es vi: eI Em pi:/}} \\

\bottomrule
\end{tabular}

\caption{Entity-level pronunciation comparison under matching and non-matching reference conditioning.}
\label{tab:special_words}
\end{table}

\section{Conclusion}

We presented KIT’s system for the IWSLT 2026 Cross-Lingual Voice Cloning task, focusing on improving multilingual in-context TTS through lightweight input conditioning and reinforcement learning fine-tuning. Starting from a strong pretrained multilingual TTS model, we investigated the effects of explicit language tags and RL-based adaptation using GRPO.

Our results show that language tags help reduce cross-lingual pronunciation drift and improve target-language consistency, while RL fine-tuning further stabilizes performance across Arabic, French, and Chinese without degrading speaker similarity or perceptual quality. In addition, we analyzed pronunciation behavior under different reference conditions and found that lexical grounding plays an important role in correctly rendering domain-specific terms and acronyms, while speaker identity remains largely preserved.

Overall, our study highlights that simple conditioning strategies combined with targeted optimization can effectively improve robustness in cross-lingual voice cloning, particularly in handling pronunciation consistency and reference-induced bias.

\section*{Acknowledgments}
This research is supported by the European Union’s Horizon Europe programme grant agreement No. 101213369 (DVPS) and KIT Campus Transfer GmbH (KCT) in accordance with the collaboration with Carnegie-AI.

\bibliography{custom}

\appendix

\end{document}